\documentclass{article}

\usepackage{microtype}
\usepackage{graphicx}
\usepackage{subfigure}
\usepackage{booktabs} 

\usepackage{hyperref}

\usepackage[accepted]{icml2025}

\usepackage{amsmath}
\usepackage{amssymb}
\usepackage{mathtools}
\usepackage{amsthm}

\usepackage[capitalize,noabbrev]{cleveref}

\theoremstyle{plain}
\newtheorem{theorem}{Theorem}[section]

\theoremstyle{definition}

\newtheorem{assumption}[theorem]{Assumption}
\theoremstyle{remark}

\usepackage[textsize=tiny]{todonotes}

\usepackage{xspace}

\definecolor{c1}{HTML}{ef78a2}

\icmltitlerunning{Collaborative Editable Model}

\begin{document}

\twocolumn[
\icmltitle{Collaborative Editable Model}



\icmlsetsymbol{equal}{*}

\begin{icmlauthorlist}
\icmlauthor{Kaiwen Tang}{equal,nus}
\icmlauthor{Aitong Wu}{equal,vak}
\icmlauthor{Yao Lu}{nus,vak}
\icmlauthor{Guangda Sun}{nus,vak}
\end{icmlauthorlist}

\icmlaffiliation{nus}{Department of Computer Science, National University of Singapore, Singapore, Singapore}
\icmlaffiliation{vak}{Vak Labs}

\icmlcorrespondingauthor{Guangda Sun}{sung@comp.nus.edu.sg}

\icmlkeywords{Machine Learning, ICML}

\vskip 0.3in
]



\printAffiliationsAndNotice{\icmlEqualContribution} 

\begin{abstract}
Vertical-domain large language models (LLMs) play a crucial role in specialized scenarios such as finance, healthcare, and law; however, their training often relies on large-scale annotated data and substantial computational resources, impeding rapid development and continuous iteration. To address these challenges, we introduce the Collaborative Editable Model (CoEM), which constructs a candidate knowledge pool from user-contributed domain snippets, leverages interactive user–model dialogues combined with user ratings and attribution analysis to pinpoint high-value knowledge fragments, and injects these fragments via in-context prompts for lightweight domain adaptation. With high-value knowledge, the LLM can generate more accurate and domain-specific content. In a financial information scenario, we collect 15k feedback from about 120 users and validate CoEM with user ratings to assess the quality of generated insights, demonstrating significant improvements in domain-specific generation while avoiding the time and compute overhead of traditional fine-tuning workflows.
\end{abstract}

\section{Introduction}

\begin{figure}[hbt!]
    \centering
    \includegraphics[width=0.9\linewidth]{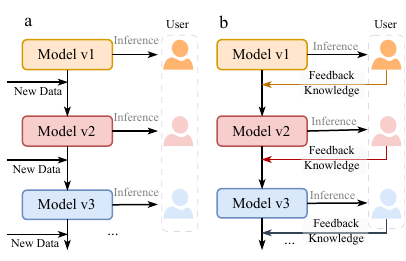}
    \caption{Comparison of Traditional Model Updating vs. Collaborative User-driven Updating. (a) Traditional model updates rely on externally provided training data; (b) Collaborative updates leverage user contributions to refine the model.}
    \label{fig:concept}
\end{figure}

Large Language Models (LLMs) have demonstrated exceptional performance in general-purpose applications and have achieved notable success in \emph{vertical domains}, such as finance, healthcare, law, and scientific research, by providing precise domain knowledge and specialized text generation~\cite{li2023large, ren2025towards, lin2025healthgpt, li2024legalagentbench}. However, the development of such vertical-domain models typically depends on large-scale annotated datasets and substantial computational resources~\cite{wu2023precedent, gururangan2020don}.
This dependency results in prolonged development cycles and slow iteration rates, making it challenging to meet the rapidly evolving demands of real-world applications.

Traditional domain adaptation methods have significant drawbacks. Supervised fine-tuning requires expensive annotation efforts and large labeled corpora, while retrieval-augmented generation (RAG)~\cite{siriwardhana2023improving} leverages external knowledge sources but depends on comprehensive, up-to-date knowledge bases~\cite{zhang2024scaling, zheng2024fine, susnjak2025automating}. These approaches rely heavily on external, annotated inputs -- data acquisition and labeling thus remain time-consuming and costly, and the limited volume of annotated domain examples constrains the benefits of scaling laws as model capacity grows.

\begin{figure*}[h]
    \centering
    \includegraphics[width=1\linewidth]{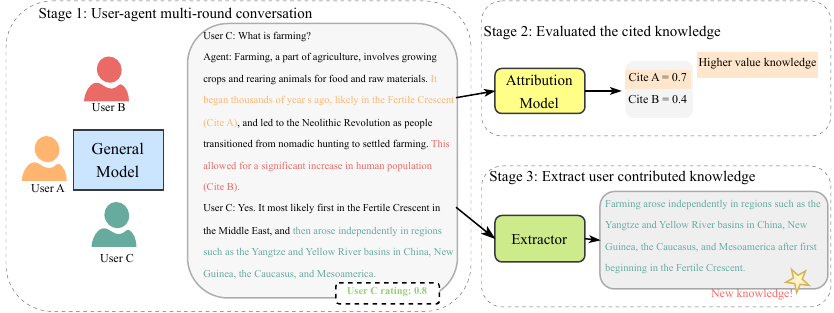}
    \caption{Overview of the Collaborative Editable Model's workflow, illustrating how user contributions drive model adaptation through multi-turn interactions and an attribution mechanism. User identities remain confidential to protect privacy during citation in generated text. The general model will be continuously updated with high-value knowledge.}
    \label{fig:arch}
\end{figure*}

Given these challenges, we explore the possibility of developing high-quality vertical-domain models without relying on manually annotated external data. Collaborative knowledge platforms, such as wiki projects, continuously evolve through user contributions, revealing the potential of crowd-sourced expertise as an ongoing learning signal~\cite{ebersbach2008wiki}. This suggests that user–model conversations can serve as the universal interface to both utilize and contribute to the model, transforming them into a valuable internal source of domain knowledge.The concept difference is shown in \cref{fig:concept}. Although a single user session may combine seeking assistance and contributing, and lack the structure and quality of traditional annotations, these conversations inherently encode valuable insights shared during interactions~\cite{chen2024large}. These insights can be extracted, evaluated, and leveraged to improve the model. However, current LLMs treat each conversation independently, without a mechanism for reintegrating conversational data into subsequent learning.




To address this gap, we present the \emph{Collaborative Editable Model (CoEM)}, a framework designed to enable the continuous improvement of vertical-domain models through user–model interactions. CoEM aggregates domain-relevant information fragments from user-contributed snippets and conversation exchanges, creating a continuously evolving knowledge pool. Users then engage in iterative dialogue sessions with the LLM, providing ratings on each generated response or insight. As the user ratings are for the whole responses, we apply attribution analysis to measure each input fragment’s contribution to overall performance, automatically identifying high-value knowledge. The fragments with high value after several iterations will be dynamically updated into the model, allowing for rapid domain adaptation without the need for extensive fine-tuning or external data. At the same time, the knowledge pool is still updating via extracting new information from user-model dialogues. The overview of the CoEM workflow is shown in \cref{fig:arch}.

CoEM offers several key advantages. First, it is \textbf{cost-efficient}, eliminating the need for costly, manually annotated data by relying on user-generated content and interactions, which significantly reduces data acquisition costs. Second, CoEM is \textbf{user-driven and responsive} to feedback. End users can identify issues or gaps in the model's performance and provide direct feedback, enabling real-time adjustments and ensuring that the system remains relevant and up-to-date. Finally, CoEM is \textbf{parallelized and scalable}. As more users engage with the system, the model benefits from an increasing volume of valuable contributions, supporting rapid updates and continuous improvement. 

In this paper, we apply CoEM to the financial domain for initial validation. We collect a set of financial news articles to construct the initial knowledge pool. A general-purpose LLM is used to generate summaries with insights for several relevant news pieces. These summaries are then presented to users for feedback. By applying attribution analysis to the user ratings, we assess the value of each knowledge fragment based on its contribution to the generated insights. These value scores are then compared to those obtained from a state-of-the-art (SOTA) financial LLM~\cite{liu2023fingpt} to validate the reliability of our data and the effectiveness of our method. Through this process, we demonstrate CoEM’s ability to incorporate valuable user-generated knowledge into the model’s context, facilitating rapid domain-specific refinement.

Our contributions are summarized as follows: 

\begin{itemize}
\vspace{-5pt}
\item We introduce the CoEM, a framework that enables the continuous improvement of vertical-domain models through user–model interactions, eliminating the need for large-scale annotated data.
\item Through the innovative contribution attribution mechanism, we address the technical challenges of extracting valuable knowledge from unstructured user dialogues, enabling rapid domain adaptation for the model.
\item The effectiveness of the CoEM framework is initially validated through experiments in the financial news domain, with future work focusing on enhancing the model’s ability to learn from user contributions through model editing or reinforcement learning.
\end{itemize}


\section{Related Work}
\subsection{Domain Adaptation for LLMs}
Early efforts in vertical-domain adaptation have predominantly relied on large-scale supervised fine-tuning. Researchers typically continue pre-training or fine-tuning a general-purpose LLM on domain-specific corpora to achieve specialized performance~\cite{gu2021domain, que2024d}; however, this approach still demands substantial annotated data and computational resources. To reduce these costs, recent work has introduced lightweight techniques, such as low-rank adaptation (LoRA)~\cite{hu2022lora, mao2025survey}, adapter modules~\cite{li2022domain}, and prompt tuning~\cite{ge2023domain, bai2024prompt, fahes2023poda}, which update only a small set of additional parameters or employ plug-and-play adaptation layers, leaving most pretrained weights untouched, and enabling rapid convergence in new domains. Retrieval-augmented generation frameworks~\cite{guu2020retrieval} further enhance domain performance by integrating external knowledge retrieval into the generation process, dynamically incorporating up-to-date documents for tasks like question answering and summarization, though they still face challenges in retriever maintenance and latency.

\subsection{Interactive Editable Models}
As human–machine collaboration paradigms evolve, editable and updatable language models have emerged as a key research direction. One strand of work employs reinforcement learning from human feedback (RLHF), using explicit user ratings or click behaviors to refine the model’s generative policies and improve alignment.~\cite{bai2022training, kaufmann2023survey}
Another focuses on fragment-level knowledge injection~\cite{czekalski2024efficiently, song2025injecting}. These methods identify “high-value” text fragments that most strongly influence outputs and prioritize their reuse. 
At the same time, crowdsourced snippets provide a flexible and rich source of domain knowledge, and by borrowing ideas from collaborative filtering, preferences can propagate across multiple agents to enable multi-directional preference diffusion. 
Although existing editable-model research has demonstrated local weight modifications and fact updates, it has yet to integrate explicit user feedback, fragment selection, and multi-agent collaborative learning into a unified framework.~\cite{casper2023open} 
In contrast, CoEM orchestrates multi-turn user–model dialogues with rating feedback, enabling lightweight, sustainable domain iteration and real-time knowledge updating.

\section{Method}

\subsection{Problem Definition}
We begin with two models: a pretrained model \( \mathcal{M}_p \), which is a general-purpose model without domain-specific adaptation, and a vertical domain model \( \mathcal{M}_v \), which represents a model tailored to a specific vertical domain. \( \mathcal{M}_p \) predicts the distribution \( P_p \) of tokens based on a given sequence \( \mathbf{s} \):
\begin{equation}
    \mathcal{M}_p(\mathbf{s}) = P_p(t | \mathbf{s})
\end{equation}

While \( \mathcal{M}_v \) predicts the domain-specific distribution \( P_v \) based on the same sequence:
\begin{equation}
\mathcal{M}_v(\mathbf{s}) = P_v(t | \mathbf{s})
\end{equation}
As we want to get a domain-specific model efficiently from the general model, our objective is to minimize the difference between the predictions of these two models by aligning \( \mathcal{M}_p \) with \( \mathcal{M}_v \):
\begin{equation}
\mathcal{L} = \min_{\mathbf{s}} \left| P_p(t | \mathbf{s}) - P_v(t | \mathbf{s}) \right|
\end{equation}
In traditional fine-tuning approaches, \( P_v \) is explicit, requiring annotated training data to guide the model. However, in our approach, \( P_v \) is implicit, as we do not rely on manually annotated data or pre-built knowledge bases. Instead, we rely on the vast amount of user feedback and contributions to guide the model’s learning direction.

\subsection{User Contribution Modeling}
\label{sec3.2}
In our framework, user contributions are categorized into two complementary forms: user feedback and user knowledge. Specifically, for model-user dialogue session \( \mathbf{d} \), \textbf{User feedback contribution} quantifies the user's evaluation of the model's output quality and domain relevance, typically expressed as a scalar score \( r \in [0,1] \). This rating indicates the relevance and usefulness of the generated content with respect to the target domain. This feedback score serves as an indicator of the effectiveness of the supporting knowledge. 
\textbf{User knowledge contribution} represents the domain-relevant information fragments extracted from the user's input during the session, each associated with a value score \( v_i \in [0,1] \).

\begin{assumption}
\label{assum}
We assume that the user feedback score \( r \in [0, 1] \) serves as a soft reinforcement learning reward signal for a model output \( \mathbf{d}_m \), reflecting the similarity between the predicted token distribution \( P_p(t|\mathbf{s}) \) and the target vertical domain distribution \( P_v(t|\mathbf{s}) \). Specifically,
\[
r = \mathcal{R}(\mathbf{d}_m) \quad \text{with} \quad r = 1 \iff P_p(t|\mathbf{s}) = P_v(t|\mathbf{s}),
\]
Moreover, for any \( r_1 > r_2 \), it holds that 
\[
\rho\big(P_p^{(r_1)}(t|\mathbf{s}), P_v(t|\mathbf{s})\big) < \rho\big(P_p^{(r_2)}(t|\mathbf{s}), P_v(t|\mathbf{s})\big),
\]
where \( \mathcal{R}(\cdot) \) denotes the reward function induced by user feedback, and \( \rho(\cdot, \cdot) \) denotes a distance metric between distributions. Thus, higher feedback corresponds to model outputs closer to the target domain distribution.
\end{assumption}

The overall objective is to guide model adaptation by aligning the weighted sum of user knowledge contributions with the vertical domain distribution:
\begin{equation}
P_v(t | \mathbf{s}) = \sum_i v_i P_i(t | \mathbf{s})
\end{equation}
This formalization enables the framework to utilize user-generated content and real-time feedback as continuous learning signals. Over successive interactions, the model incrementally evolves to better capture domain-specific knowledge, eliminating the need for extensive manually annotated data and facilitating rapid adaptation within vertical domains.

\subsection{CoEM Knowledge Pool}


Consider a multi-round dialogue session \( \mathbf{d} \) between the model and the user. During this session, the model generated output \( \mathbf{d}_m \) is supported by a collection of domain-specific knowledge fragments \( \{ k_1, k_2, \dots \} \) retrieved from the current knowledge pool \( \mathbf{K} \). Each knowledge fragment \( k_i \) is associated with a value score \( v_i \), representing its estimated relevance and utility within the vertical domain.

After receiving the model's response, the user provides the feedback score \(r\in [0,1] \) as introduced in \cref{sec3.2}.
To quantify the contribution of individual knowledge fragments to the overall feedback, an attributor \( \mathcal{A} \) is employed to assign a contribution weight \( p_i \) to each fragment \( k_i \). The adjusted contribution value for fragment \( k_i \) in the current session is then computed as
\begin{equation}
\label{score'}
v_i' = p_i \cdot r.
\end{equation}
Treating each dialogue session as an iteration, the value scores of the knowledge fragments are updated according to an exponential moving average scheme~\cite{lucas1990exponentially}:
\begin{equation}
\label{score}
v_i \leftarrow (1 - \alpha) v_i + \alpha v_i',
\end{equation}
Where \( \alpha \in (0,1) \) denotes the learning rate that controls the magnitude of the update. Knowledge fragments whose value scores fall below a predefined threshold \( \theta \) after \( n \) iterations are pruned from the knowledge pool, as they are considered insufficiently relevant and useful to the domain.

Concurrently, the user’s input during the dialogue session, denoted \( \mathbf{d}_u \), is processed by a knowledge extractor \( \mathcal{E} \) to identify potential new domain-relevant knowledge fragments \( \{ u_1, u_2, \dots \} \). These newly extracted fragments are incorporated into the knowledge pool \( \mathbf{K} \), thereby expanding the pool. 

We initialize newly added knowledge fragments with a value score of 1, reflecting an \emph{optimistic initialization} strategy~\cite{lobel2022optimistic}. This approach encourages the model to treat new knowledge as valuable initially, while the exponential moving average update bounds the scores below 1 and allows subsequent feedback to adjust them dynamically. Optimistic initialization is commonly used in reinforcement and online learning to promote exploration and avoid premature dismissal of novel information. The whole process of updating the CoEM knowledge pool is shown in \cref{alg1}.
\begin{algorithm}
\caption{CoEM Knowledge Pool Update}
\label{alg1}
\begin{algorithmic}[1]
\STATE \textbf{Input:} 
General model $\mathcal{M}_p$, Knowledge pool \( \mathbf{K} = \{(k_i, v_i)\} \), dialogue session \(\mathbf{d} = (\mathbf{d}_m, \mathbf{d}_u)\), 
learning rate \(\alpha\), value threshold \(\theta\), attribution method \(\mathcal{A}\), knowledge extractor \(\mathcal{E}\).

\STATE \textbf{Output:} Updated knowledge pool \( \mathbf{K} \)

\STATE $ \mathbf{d}_m  \gets \mathcal{M}_p(\hat{K}),  \hat{K} \subseteq \mathbf{K} $

\STATE $r \gets \mathcal{R}(\mathbf{d}_m), r \in [0,1]$ \{Get user feedback on $ \mathbf{d}_m$\}

\vspace{3pt}
\STATE \textbf{Attributor:}
\STATE ${p_i} = \mathcal{A}(\mathbf{d}_m, k_i), k_i \in \hat{K}$ 

\FOR{ \( k_i \in \hat{K} \)}
    \STATE $v_i \gets (1 - \alpha) \cdot v_i + \alpha \cdot p_i \cdot r$
\ENDFOR

\vspace{3pt}
\STATE \textbf{Extractor:}
\STATE ${u_j} = \mathcal{E}(\mathbf{d}_u)$

\STATE  \( v_{u_j} \gets 1 \quad \forall u_j \) \{Initialize value scores.\}
\STATE $\mathbf{K} \gets \mathbf{K} \cup \{(u_j, v_{u_j})\}$

\STATE $\mathbf{K} \leftarrow \mathbf{K} \setminus \{ k_i \mid v_i < \theta \}$

\STATE \textbf{Return} updated knowledge pool \( \mathbf{K} \)
\end{algorithmic}
\end{algorithm}

Through this iterative process of feedback-driven value updating and continuous knowledge extraction, the knowledge pool is dynamically refined and enlarged. This mechanism enables the model to progressively enhance its domain expertise and improve response quality, all achieved through ongoing user interaction without reliance on extensive manual annotation.

\begin{figure*}[htbp]
  \centering
  \begin{minipage}{0.24\textwidth}
    \centering
    \caption{Attribution score distribution weighted by user feedback.}
    \label{fig:score}
    \includegraphics[width=\linewidth]{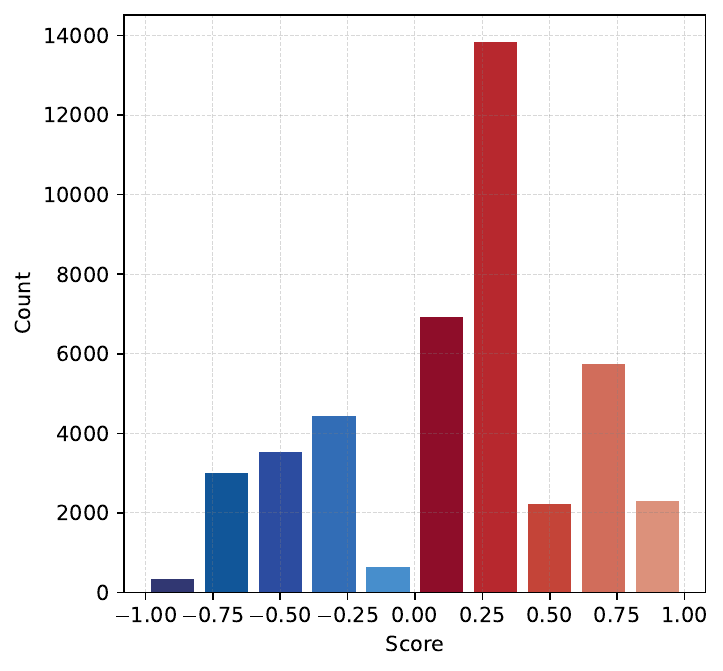}
  \end{minipage}%
  \hfill
  \begin{minipage}{0.24\textwidth}
    \caption{Distribution of knowledge fragment value scores after iterative updates.}
    \centering
    \includegraphics[width=\linewidth]{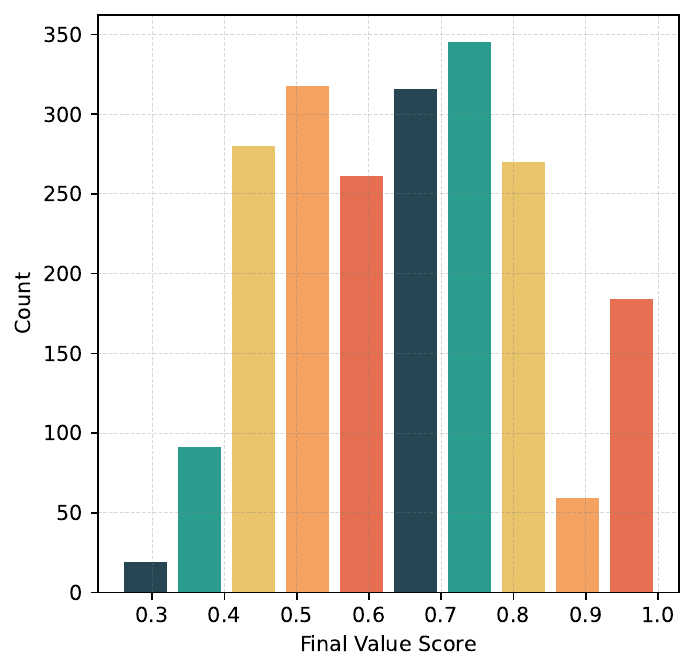}
    \label{fig:value}
  \end{minipage}%
  \hfill
  \begin{minipage}{0.24\textwidth}
    \centering
    \caption{Distribution of knowledge fragment value scores from baseline model.}
    \includegraphics[width=\linewidth]{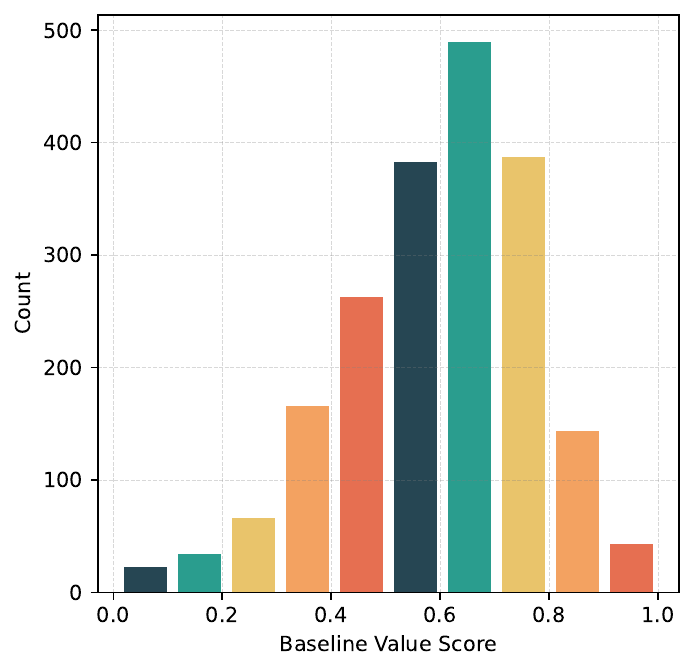}
    \label{fig:value_baseline}
  \end{minipage}
  \hfill
  \begin{minipage}{0.24\textwidth}
    \centering
    \caption{Proportion of high-value knowledge fragments under varying learning rates.}
    \label{fig:lr}
    \includegraphics[width=\linewidth]{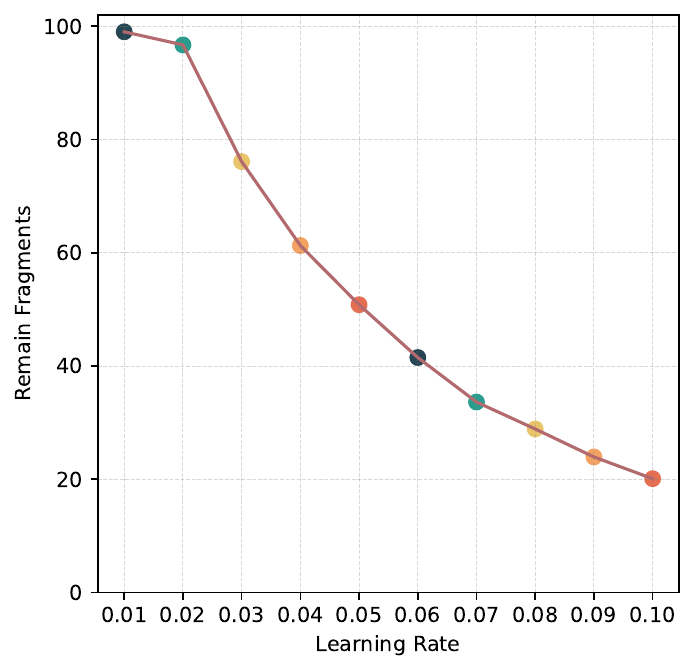}
  \end{minipage}
\end{figure*}

\subsection{Attribution Mechanism}  
In this section, we briefly introduce the design of our attribution mechanism. To quantitatively assess the value of knowledge \( k_i \) from the model generated text \( \mathbf{d}_m \) and user feedback $r$, CoEM employs an attribution function \( \mathcal{A} \) defined as
\begin{equation}
    \mathcal{A}(\mathbf{d}_m, k_i) \in [0,1]
\end{equation}

which represents the strength of the causal influence of knowledge \( k_i \) on the generated response \( \mathbf{d}_m \). Intuitively, this value measures how much the distribution of \( \mathbf{d}_m \) would differ if the model had not incorporated knowledge from \( k_i \).

The attribution model satisfies natural boundary conditions:
\begin{equation}
\mathcal{A}(k_i, k_i) = 1
\end{equation}
Notably, \( \mathcal{A} \) operates without requiring domain-specific knowledge distribution \( P_v \), ensuring broad applicability.

Considering a set of model outputs \( \{ \mathbf{d}_m^i \} \) generated during interactions and corresponding user feedback scores \( \{ r_i \} \in [0,1] \), the value score \( v_j \) for each input knowledge \( u_j \) is computed as the expected weighted attribution:
\begin{equation}
    v_j = \mathbb{E}_{\mathbf{d}_m^i} \left[ \mathcal{A}(\mathbf{d}_m^i, u_j) \cdot r_i \right]
\end{equation}

This expectation aggregates the causal effect of \( u_j \) across multiple model responses, modulated by user feedback, yielding a statistically meaningful measure of the input's overall contribution.


\section{Experiment}
In this work, we performed experiments in the financial domain as a starting point. We used the Gemini-2.0-flash-lite API to access Gemini~\cite{team2023gemini} as the general model. For the vertical domain model, we employed FinGPT~\cite{liu2023fingpt} from Huggingface\footnote{\url{https://huggingface.co/FinGPT/fingpt-mt_llama3-8b_lora}} built based on LLaMa-3~\cite{grattafiori2024llama}.

\subsection{Data Construction}
We first collected 2,578 knowledge fragments related to finance and cryptocurrency news from a variety of news websites and forums, including Yahoo Finance, MarketWatch.com, Cointelegraph.com, Crypto.news, and so on, in both English and Chinese. These fragments served as the initial knowledge pool for our model. 

We then prompted the general model to generate summaries with viewpoints for 3 related fragments in the pool. These summaries were presented to users, who provided feedback in the form of ``like" or ``dislike" evaluations, which serve as user feedback contributions. The users involved in the feedback process all had relevant knowledge and backgrounds in the finance and cryptocurrency domains. We received a total of 15,040 feedback responses from approximately 120 users, with a "like" to "dislike" ratio of 10,696: 4,344. Importantly, all user data was anonymized, and no personal information was recorded, ensuring the privacy of the contributors.

Then we use another general model as the attributor. Given the model-generated summary with viewpoints and user feedback, the attributor model evaluates the contribution of each news article to the generated summary. The feedback score provided by the user is used to compute the contribution weight for each news article with \cref{score}. As we only collect the user rating as likes or dislikes, we assign $r=1$ to what the user likes and $-1$ to what the user dislikes. We choose the learning rate $\alpha$ as 0.03.

\subsection{Results}
\label{res}
\paragraph{High-value knowledge}
The key step for CoEM is to distinguish which contributions are useful to the model.
After the iterative updating of the value score, we regard the knowledge with high scores as valuable knowledge for the model to learning to become a vertical domain model. In the current dataset, after calculating with \cref{score'}, we got the value score for each iteration. The distribution is shown as \cref{fig:score}.

With the value score $v'$ for each iteration, we update the value score for each knowledge fragment in the knowledge pool; the distribution of the total value score is shown in \cref{fig:value}. Only 0.14\% of the fragments didn't get feedback from the users, with a learning rate of 0.03 and a threshold of 0.5, 76.11\% of the fragments remain in the knowledge pool and are marked as high-value knowledge. 

\paragraph{Comparison with Vertical Domain Model}
To illustrate the effectiveness of our user contribution strategy, we also used the vertical domain model to evaluate our knowledge and validate our \cref{assum}. Specifically, we ask the vertical domain model to rate the relevance and usefulness of each knowledge fragment and compare the results with the value scores in our knowledge pool. The distribution is shown in \cref{fig:value_baseline}. From FinGPT, 73.62\% of the fragments are marked as high-value knowledge. Among all the high-value knowledge in the knowledge pool, 76.11\% of them are also recognized by the baseline model, indicating the effectiveness of our methods. 

\paragraph{Learning Rate of CoEM}
We also performed an ablation study on the learning rate to illustrate the reason for picking 0.03 in the experiment. As shown in \cref{fig:lr}, when the learning rate increases, the proportion of high-value knowledge decreases, so 0.03 could serve as a value to keep enough knowledge in the pool. This may also show that our attributor tends to give lower marks, indicating that the attributor and the threshold require further exploration.


\section{Discussion}
 
\paragraph{User Feedback and Attribution Mechanism}  

CoEM heavily depends on integrating user feedback with the attribution model. However, accurately attributing individual user contributions to the model’s output remains challenging as we discussed in \cref{res}, especially in multi-turn dialogues where prior context and existing knowledge interact in complex ways. One possible approach to address this challenge is using Shapley value-based attribution, which offers a principled and fair method for quantifying the contribution of each user input, enhancing transparency. Additionally, as shown in \cref{fig:lr}, the current optimistic initialization strategy may lead to a decrease in attribution scores over time. To mitigate this, future versions of CoEM may incorporate a decay mechanism within the value score update process to better balance score evolution.

\paragraph{Scalability and Further Experiment}  
As CoEM relies on continuous user interaction, scalability is a key concern. With an increasing number of users, the computational overhead for real-time updates and knowledge pool management could become a bottleneck. One possible solution could be to implement incremental learning techniques that allow for more efficient adaptation with minimal computational cost. For instance, leveraging sparse updates rather than full model retraining can drastically reduce resource consumption while preserving model performance. Additionally, the use of distributed learning frameworks can help balance the load across different computing nodes, ensuring scalability without compromising real-time performance. Optimizing these algorithms will be crucial for large-scale deployment.

\paragraph{User Privacy and Data Security}  
A major advantage of CoEM is its inherent respect for user privacy. 
Unlike traditional models, CoEM eliminates the need for centralized data collection, directly addressing privacy concerns arising from large-scale data usage. However, to ensure the protection of the user's privacy, we need to design the decentralized learning process to keep the user's data local. A possible solution could be integrating differential privacy techniques~\cite{yan2024protecting}, which add noise to the data, ensuring that individual user contributions cannot be traced back to a specific user while still allowing the model to learn effectively from the aggregated knowledge. Additionally, secure multi-party computation (SMPC)~\cite{knott2021crypten} could be utilized to allow the model to learn from user data without directly accessing sensitive information. These techniques would further reinforce CoEM’s privacy-preserving framework, ensuring robust protection for users while maintaining high model performance.

\section{Conclusion}

We propose the Collaborative Editable Model (CoEM), a novel user-driven framework for efficiently adapting large language models to vertical domains. CoEM maintains a dynamic knowledge pool updated through user contributions and multi-turn dialogue feedback. A core component is the attribution mechanism, which quantifies the impact of each knowledge fragment on model outputs, enabling precise updates to their value scores without relying on costly annotated datasets or extensive fine-tuning.

Experimental results on financial news demonstrate CoEM’s effectiveness in identifying and reinforcing high-value domain knowledge, with over 76\% agreement with a state-of-the-art vertical domain model. We also study the influence of learning rate on knowledge retention, revealing important trade-offs for practical deployment. Overall, CoEM offers a scalable, privacy-aware, and interactive approach for continuous vertical domain adaptation driven by collaborative user feedback. 

\section*{Impact Statement}

This paper presents work whose goal is to advance the field of 
Machine Learning. There are many potential societal consequences 
of our work, none of which we feel must be specifically highlighted here.


\bibliography{example_paper}

\begin{thebibliography}{33}
\providecommand{\natexlab}[1]{#1}
\providecommand{\url}[1]{\texttt{#1}}
\expandafter\ifx\csname urlstyle\endcsname\relax
  \providecommand{\doi}[1]{doi: #1}\else
  \providecommand{\doi}{doi: \begingroup \urlstyle{rm}\Url}\fi

\bibitem[Bai et~al.(2024)Bai, Zhang, Zhou, Huang, Luan, Wang, and Chen]{bai2024prompt}
Bai, S., Zhang, M., Zhou, W., Huang, S., Luan, Z., Wang, D., and Chen, B.
\newblock Prompt-based distribution alignment for unsupervised domain adaptation.
\newblock In \emph{Proceedings of the AAAI conference on artificial intelligence}, volume~38, pp.\  729--737, 2024.

\bibitem[Bai et~al.(2022)Bai, Jones, Ndousse, Askell, Chen, DasSarma, Drain, Fort, Ganguli, Henighan, et~al.]{bai2022training}
Bai, Y., Jones, A., Ndousse, K., Askell, A., Chen, A., DasSarma, N., Drain, D., Fort, S., Ganguli, D., Henighan, T., et~al.
\newblock Training a helpful and harmless assistant with reinforcement learning from human feedback.
\newblock \emph{arXiv preprint arXiv:2204.05862}, 2022.

\bibitem[Casper et~al.(2023)Casper, Davies, Shi, Gilbert, Scheurer, Rando, Freedman, Korbak, Lindner, Freire, et~al.]{casper2023open}
Casper, S., Davies, X., Shi, C., Gilbert, T.~K., Scheurer, J., Rando, J., Freedman, R., Korbak, T., Lindner, D., Freire, P., et~al.
\newblock Open problems and fundamental limitations of reinforcement learning from human feedback.
\newblock \emph{arXiv preprint arXiv:2307.15217}, 2023.

\bibitem[Chen et~al.(2024)Chen, Liu, Huang, Wu, Liu, Jiang, Pu, Lei, Chen, Wang, et~al.]{chen2024large}
Chen, J., Liu, Z., Huang, X., Wu, C., Liu, Q., Jiang, G., Pu, Y., Lei, Y., Chen, X., Wang, X., et~al.
\newblock When large language models meet personalization: Perspectives of challenges and opportunities.
\newblock \emph{World Wide Web}, 27\penalty0 (4):\penalty0 42, 2024.

\bibitem[Czekalski \& Watson(2024)Czekalski and Watson]{czekalski2024efficiently}
Czekalski, E. and Watson, D.
\newblock Efficiently updating domain knowledge in large language models: Techniques for knowledge injection without comprehensive retraining.
\newblock 2024.

\bibitem[Ebersbach et~al.(2008)Ebersbach, Glaser, Heigl, and Warta]{ebersbach2008wiki}
Ebersbach, A., Glaser, M., Heigl, R., and Warta, A.
\newblock \emph{Wiki: web collaboration}.
\newblock springer science \& business media, 2008.

\bibitem[Fahes et~al.(2023)Fahes, Vu, Bursuc, P{\'e}rez, and De~Charette]{fahes2023poda}
Fahes, M., Vu, T.-H., Bursuc, A., P{\'e}rez, P., and De~Charette, R.
\newblock Poda: Prompt-driven zero-shot domain adaptation.
\newblock In \emph{Proceedings of the IEEE/CVF International Conference on Computer Vision}, pp.\  18623--18633, 2023.

\bibitem[Ge et~al.(2023)Ge, Huang, Xie, Lai, Song, Li, and Huang]{ge2023domain}
Ge, C., Huang, R., Xie, M., Lai, Z., Song, S., Li, S., and Huang, G.
\newblock Domain adaptation via prompt learning.
\newblock \emph{IEEE Transactions on Neural Networks and Learning Systems}, 2023.

\bibitem[Grattafiori et~al.(2024)Grattafiori, Dubey, Jauhri, Pandey, Kadian, Al-Dahle, Letman, Mathur, Schelten, Vaughan, et~al.]{grattafiori2024llama}
Grattafiori, A., Dubey, A., Jauhri, A., Pandey, A., Kadian, A., Al-Dahle, A., Letman, A., Mathur, A., Schelten, A., Vaughan, A., et~al.
\newblock The llama 3 herd of models.
\newblock \emph{arXiv preprint arXiv:2407.21783}, 2024.

\bibitem[Gu et~al.(2021)Gu, Tinn, Cheng, Lucas, Usuyama, Liu, Naumann, Gao, and Poon]{gu2021domain}
Gu, Y., Tinn, R., Cheng, H., Lucas, M., Usuyama, N., Liu, X., Naumann, T., Gao, J., and Poon, H.
\newblock Domain-specific language model pretraining for biomedical natural language processing.
\newblock \emph{ACM Transactions on Computing for Healthcare (HEALTH)}, 3\penalty0 (1):\penalty0 1--23, 2021.

\bibitem[Gururangan et~al.(2020)Gururangan, Marasovi{\'c}, Swayamdipta, Lo, Beltagy, Downey, and Smith]{gururangan2020don}
Gururangan, S., Marasovi{\'c}, A., Swayamdipta, S., Lo, K., Beltagy, I., Downey, D., and Smith, N.~A.
\newblock Don’t stop pretraining: Adapt language models to domains and tasks.
\newblock In \emph{Proceedings of the 58th Annual Meeting of the Association for Computational Linguistics}, pp.\  8342--8360, 2020.

\bibitem[Guu et~al.(2020)Guu, Lee, Tung, Pasupat, and Chang]{guu2020retrieval}
Guu, K., Lee, K., Tung, Z., Pasupat, P., and Chang, M.
\newblock Retrieval augmented language model pre-training.
\newblock In \emph{International conference on machine learning}, pp.\  3929--3938. PMLR, 2020.

\bibitem[Hu et~al.(2022)Hu, Shen, Wallis, Allen-Zhu, Li, Wang, Wang, Chen, et~al.]{hu2022lora}
Hu, E.~J., Shen, Y., Wallis, P., Allen-Zhu, Z., Li, Y., Wang, S., Wang, L., Chen, W., et~al.
\newblock Lora: Low-rank adaptation of large language models.
\newblock \emph{ICLR}, 1\penalty0 (2):\penalty0 3, 2022.

\bibitem[Kaufmann et~al.(2023)Kaufmann, Weng, Bengs, and H{\"u}llermeier]{kaufmann2023survey}
Kaufmann, T., Weng, P., Bengs, V., and H{\"u}llermeier, E.
\newblock A survey of reinforcement learning from human feedback.
\newblock \emph{arXiv preprint arXiv:2312.14925}, 10, 2023.

\bibitem[Knott et~al.(2021)Knott, Venkataraman, Hannun, Sengupta, Ibrahim, and van~der Maaten]{knott2021crypten}
Knott, B., Venkataraman, S., Hannun, A., Sengupta, S., Ibrahim, M., and van~der Maaten, L.
\newblock Crypten: Secure multi-party computation meets machine learning.
\newblock \emph{Advances in Neural Information Processing Systems}, 34:\penalty0 4961--4973, 2021.

\bibitem[Li et~al.(2024)Li, Chen, Yang, Ai, Jia, Liu, Lin, Wu, Yuan, Hu, et~al.]{li2024legalagentbench}
Li, H., Chen, J., Yang, J., Ai, Q., Jia, W., Liu, Y., Lin, K., Wu, Y., Yuan, G., Hu, Y., et~al.
\newblock Legalagentbench: Evaluating llm agents in legal domain.
\newblock \emph{arXiv preprint arXiv:2412.17259}, 2024.

\bibitem[Li et~al.(2023)Li, Wang, Ding, and Chen]{li2023large}
Li, Y., Wang, S., Ding, H., and Chen, H.
\newblock Large language models in finance: A survey.
\newblock In \emph{Proceedings of the fourth ACM international conference on AI in finance}, pp.\  374--382, 2023.

\bibitem[Li et~al.(2022)Li, Ren, Jiang, Li, Zhang, and Li]{li2022domain}
Li, Z., Ren, K., Jiang, X., Li, B., Zhang, H., and Li, D.
\newblock Domain generalization using pretrained models without fine-tuning.
\newblock \emph{arXiv preprint arXiv:2203.04600}, 2022.

\bibitem[Lin et~al.(2025)Lin, Zhang, Li, Yuan, Yu, Li, He, Jiang, Li, Song, et~al.]{lin2025healthgpt}
Lin, T., Zhang, W., Li, S., Yuan, Y., Yu, B., Li, H., He, W., Jiang, H., Li, M., Song, X., et~al.
\newblock Healthgpt: A medical large vision-language model for unifying comprehension and generation via heterogeneous knowledge adaptation.
\newblock \emph{arXiv preprint arXiv:2502.09838}, 2025.

\bibitem[Liu et~al.(2023)Liu, Wang, Yang, and Zha]{liu2023fingpt}
Liu, X.-Y., Wang, G., Yang, H., and Zha, D.
\newblock Fingpt: Democratizing internet-scale data for financial large language models.
\newblock \emph{arXiv preprint arXiv:2307.10485}, 2023.

\bibitem[Lobel et~al.(2022)Lobel, Gottesman, Allen, Bagaria, and Konidaris]{lobel2022optimistic}
Lobel, S., Gottesman, O., Allen, C., Bagaria, A., and Konidaris, G.
\newblock Optimistic initialization for exploration in continuous control.
\newblock In \emph{Proceedings of the AAAI Conference on Artificial Intelligence}, volume~36, pp.\  7612--7619, 2022.

\bibitem[Lucas \& Saccucci(1990)Lucas and Saccucci]{lucas1990exponentially}
Lucas, J.~M. and Saccucci, M.~S.
\newblock Exponentially weighted moving average control schemes: properties and enhancements.
\newblock \emph{Technometrics}, 32\penalty0 (1):\penalty0 1--12, 1990.

\bibitem[Mao et~al.(2025)Mao, Ge, Fan, Xu, Mi, Hu, and Gao]{mao2025survey}
Mao, Y., Ge, Y., Fan, Y., Xu, W., Mi, Y., Hu, Z., and Gao, Y.
\newblock A survey on lora of large language models.
\newblock \emph{Frontiers of Computer Science}, 19\penalty0 (7):\penalty0 197605, 2025.

\bibitem[Que et~al.(2024)Que, Liu, Zhang, Zhang, Qu, Ma, Duan, Bai, Wang, Zhang, et~al.]{que2024d}
Que, H., Liu, J., Zhang, G., Zhang, C., Qu, X., Ma, Y., Duan, F., Bai, Z., Wang, J., Zhang, Y., et~al.
\newblock D-cpt law: Domain-specific continual pre-training scaling law for large language models.
\newblock \emph{Advances in Neural Information Processing Systems}, 37:\penalty0 90318--90354, 2024.

\bibitem[Ren et~al.(2025)Ren, Jian, Ren, Leng, Xie, and Zhang]{ren2025towards}
Ren, S., Jian, P., Ren, Z., Leng, C., Xie, C., and Zhang, J.
\newblock Towards scientific intelligence: A survey of llm-based scientific agents.
\newblock \emph{arXiv preprint arXiv:2503.24047}, 2025.

\bibitem[Siriwardhana et~al.(2023)Siriwardhana, Weerasekera, Wen, Kaluarachchi, Rana, and Nanayakkara]{siriwardhana2023improving}
Siriwardhana, S., Weerasekera, R., Wen, E., Kaluarachchi, T., Rana, R., and Nanayakkara, S.
\newblock Improving the domain adaptation of retrieval augmented generation (rag) models for open domain question answering.
\newblock \emph{Transactions of the Association for Computational Linguistics}, 11:\penalty0 1--17, 2023.

\bibitem[Song et~al.(2025)Song, Yan, Liu, Fang, Li, Yan, and Chen]{song2025injecting}
Song, Z., Yan, B., Liu, Y., Fang, M., Li, M., Yan, R., and Chen, X.
\newblock Injecting domain-specific knowledge into large language models: a comprehensive survey.
\newblock \emph{arXiv preprint arXiv:2502.10708}, 2025.

\bibitem[Susnjak et~al.(2025)Susnjak, Hwang, Reyes, Barczak, McIntosh, and Ranathunga]{susnjak2025automating}
Susnjak, T., Hwang, P., Reyes, N., Barczak, A.~L., McIntosh, T., and Ranathunga, S.
\newblock Automating research synthesis with domain-specific large language model fine-tuning.
\newblock \emph{ACM Transactions on Knowledge Discovery from Data}, 19\penalty0 (3):\penalty0 1--39, 2025.

\bibitem[Team et~al.(2023)Team, Anil, Borgeaud, Alayrac, Yu, Soricut, Schalkwyk, Dai, Hauth, Millican, et~al.]{team2023gemini}
Team, G., Anil, R., Borgeaud, S., Alayrac, J.-B., Yu, J., Soricut, R., Schalkwyk, J., Dai, A.~M., Hauth, A., Millican, K., et~al.
\newblock Gemini: a family of highly capable multimodal models.
\newblock \emph{arXiv preprint arXiv:2312.11805}, 2023.

\bibitem[Wu et~al.(2023)Wu, Zhou, Liu, Lu, Liu, Zhang, Sun, Wu, and Kuang]{wu2023precedent}
Wu, Y., Zhou, S., Liu, Y., Lu, W., Liu, X., Zhang, Y., Sun, C., Wu, F., and Kuang, K.
\newblock Precedent-enhanced legal judgment prediction with llm and domain-model collaboration.
\newblock In \emph{Proceedings of the 2023 Conference on Empirical Methods in Natural Language Processing}, pp.\  12060--12075, 2023.

\bibitem[Yan et~al.(2024)Yan, Li, Xu, Dong, Zhang, Ren, and Cheng]{yan2024protecting}
Yan, B., Li, K., Xu, M., Dong, Y., Zhang, Y., Ren, Z., and Cheng, X.
\newblock On protecting the data privacy of large language models (llms): A survey.
\newblock \emph{arXiv preprint arXiv:2403.05156}, 2024.

\bibitem[Zhang et~al.(2024)Zhang, Liu, Cherry, and Firat]{zhang2024scaling}
Zhang, B., Liu, Z., Cherry, C., and Firat, O.
\newblock When scaling meets llm finetuning: The effect of data, model and finetuning method.
\newblock \emph{arXiv preprint arXiv:2402.17193}, 2024.

\bibitem[Zheng et~al.(2024)Zheng, Hong, Liu, Wang, Su, Liang, and Wu]{zheng2024fine}
Zheng, J., Hong, H., Liu, F., Wang, X., Su, J., Liang, Y., and Wu, S.
\newblock Fine-tuning large language models for domain-specific machine translation.
\newblock \emph{arXiv preprint arXiv:2402.15061}, 2024.

\end{thebibliography}
\bibliographystyle{icml2025}

\end{document}